\documentclass[runningheads]{llncs}

\usepackage[T1]{fontenc}
\usepackage{graphicx}
\usepackage{booktabs}
\usepackage{float}
\usepackage[hyphens]{url}
\usepackage{cite}
\usepackage{amssymb}
\usepackage{placeins}
\usepackage{orcidlink}
\usepackage{capt-of}
\usepackage{marvosym}

\title{KQML-Liver: Interpretable Hybrid Quantum Machine Learning with Kolmogorov--Arnold Networks for Liver Disease Classification}
\titlerunning{KQML-Liver: Hybrid KAN--Quantum ML}

\author{
Nhat-Long Nguyen\textsuperscript{(\Letter)}\orcidlink{0009-0009-0079-967X} \and
Viet-Loi Dinh\orcidlink{0009-0006-4910-2667} \and
Trong-Nghia Pham\orcidlink{0000-0001-5182-6003}
}
\authorrunning{Nguyen et al.}
\institute{
Faculty of Information Technology, University of Science, Vietnam National University, Ho Chi Minh City, Vietnam \\
\email{\{22120194, 22120188\}@student.hcmus.edu.vn, ptnghia@fit.hcmus.edu.vn} \\
}

\begin{document}
\raggedbottom
\sloppy
\hbadness=10000
\vbadness=10000
\hfuzz=3000pt
\vfuzz=10pt
\let\balance\relax
\setlength{\abovedisplayskip}{6pt}
\setlength{\belowdisplayskip}{6pt}
\setlength{\abovedisplayshortskip}{4pt}
\setlength{\belowdisplayshortskip}{4pt}
\setlength{\parskip}{0pt}
\setlength{\parindent}{12pt}

\maketitle

\begin{abstract}

Liver disease remains a major global health concern, highlighting the need for reliable and accessible early detection methods. Although recent advances in machine learning, deep learning and quantum machine learning (QML) have introduced promising diagnostic tools, many approaches suffer from limited interpretability and inflated performance caused by data leakage or overly simplistic evaluation protocols. These constraints reduce their reliability in real-world clinical practice. This paper proposes KQML-Liver, a hybrid framework that integrates QML with Kolmogorov–Arnold Networks (KANs) to enhance predictive performance and interpretability. The architecture combines parameterized quantum circuits for expressive feature representation with KAN-based functional mappings that capture nonlinear relationships in an interpretable manner. Evaluated on the Indian Liver Patient Dataset (ILPD) under a rigorous preprocessing protocol preventing data leakage, the proposed model achieves 82\% classification accuracy, and 100\% recall, detecting all positive cases , outperforming existing baselines while providing improved transparency in clinical predictions. These results suggest that hybrid KAN--quantum architectures offer a promising direction for interpretable AI-assisted medical diagnosis.
\end{abstract}


\section{Introduction}

Liver diseases remain a major global health challenge, causing over 2 million deaths annually~\cite{ref:liverdeath2023}. Viral hepatitis is a leading contributor, claiming a life every 30 seconds worldwide according to WHO ~\cite{ref:who-hepatitis}. The asymptomatic nature of early-stage liver disease, combined with limited access to screening, delays diagnosis and timely intervention, resulting in preventable morbidity and mortality.

Early detection is therefore essential for improving treatment outcomes. However, conventional diagnostic procedures are often costly, invasive, and subject to human error, motivating the adoption of Artificial Intelligence (AI) for automated clinical decision support. In particular, machine learning (ML) has demonstrated considerable success in disease prediction and classification across various medical applications.

Despite their success, classical ML models often struggle with small or imbalanced datasets and have limited capacity to capture complex feature interactions. While deep learning approaches provide stronger representation learning, they typically require large-scale data and are prone to overfitting when applied to small, noisy, or imbalanced medical datasets. Moreover, their effectiveness on structured tabular data is often limited, and their black-box nature raises concerns regarding interpretability and trustworthiness in clinical decision-making.

Quantum Machine Learning (QML) has recently emerged as a promising paradigm combining quantum computing with classical techniques to model complex feature interactions. By exploiting superposition and entanglement, QML enables expressive, nonlinear feature mappings into high-dimensional Hilbert spaces. However, current quantum systems are constrained by noise and hardware limitations in the Noisy Intermediate-Scale Quantum (NISQ) era, restricting the scalability and practicality of purely quantum approaches.

To resolve these problems, we introduce KQML-Liver, an interpretable hybrid KAN--quantum framework for liver disease classification. Our pipeline leverages QML to encode input data into high-dimensional feature spaces, enabling richer representation learning while maintaining computational efficiency with a limited number of qubits. To ensure interpretability in clinical settings, Kolmogorov–Arnold Networks (KANs)~\cite{ref:kan2024} are integrated, providing intrinsic transparency through learnable, symbolically expressible functions. This facilitates structured analysis of feature contributions, transforming the model from a black box into a transparent and clinically interpretable decision support system.

In addition, the framework addresses data imbalance using Effective-Number-of-Samples (ENS) weighting~\cite{ref:cui2019}, improving minority class learning while preserving the original data distribution.

The main contributions of this study are summarized as follows:

\begin{enumerate}
\item We propose a novel hybrid KAN--quantum framework that leverages functional representations and high-dimensional quantum encoding to capture highly complex nonlinear relationships, opening a promising direction for tabular learning problems characterized by limited features and small datasets---particularly in high-stakes domains such as healthcare.

\item We demonstrate that the KQML-Liver achieves competitive performance on the ILPD benchmark without altering the original data distribution. 

\item We further enhance model transparency by incorporating Kolmogorov--Arnold Networks, which provide intrinsic interpretability through functional decomposition and enable the extraction of explicit symbolic expressions for clinical deployment. 

\end{enumerate}

\section{Related Work}

\subsection{Classical Machine Learning Approaches}

Early studies explored a variety of machine learning approaches for liver disease prediction. Sontakke et al.~\cite{ref:sontakke2017} compared neural networks and SVMs, while later works incorporated classifiers such as Random Forest, Decision Tree, KNN, and advanced SVM variants. Azam et al.~\cite{ref:azam2020} provided a comprehensive evaluation of these methods, and Kumar and Thakur~\cite{ref:kumarthakur2021} addressed class imbalance using a Variable-Neighbor Weighted Fuzzy KNN to improve minority-class recognition.

In parallel, data-centric approaches have demonstrated the effectiveness of more advanced preprocessing pipelines can lead to significant performance gains. Quadir et al.~\cite{ref:quadir2023} combined multivariate imputation, skewness reduction, upsampling, and feature scaling with ensemble learning methods, yielding strong classification results. Similarly, Amin et al.~\cite{ref:amin2023} integrated multiple statistical techniques, including PCA, Factor Analysis, and Linear Discriminant Analysis, into a unified feature representation that enhances downstream classification performance.

Nevertheless, these approaches largely treat model design and data preprocessing as separate optimization steps, while remaining sensitive to class imbalance and offering limited intrinsic interpretability for clinical use.

\subsection{Quantum Machine Learning Approaches}

Recent studies have explored QML as a promising direction for improving liver disease classification by leveraging high-dimensional quantum representations and hybrid learning frameworks.

Raubitzek et al.~\cite{ref:raubitzek2024} extended quantum kernel methods using Lie-group symmetries to construct structured feature maps, demonstrating how symmetry-aware encoding can enhance representational expressiveness. Safriandono et al.~\cite{ref:safriandono2024} designed a quantum feature engineering (QFE) framework that maps classical data into quantum states using parameterized circuits composed of Hadamard, rotation, and entangling gates. Their results indicate that QFE can improve the performance of downstream classical classifiers, particularly when combined with preprocessing strategies such as SMOTE-Tomek.

Hybrid quantum--classical models have also been investigated to balance expressiveness and practical feasibility. Donaire et al.~\cite{ref:qml2025} introduced a hybrid neural architecture that combines classical dense layers with a quantum circuit. Bhaskaran and Prasanna~\cite{ref:bhaskaran2024} explored quantum-enhanced KNN (QKNN) with different distance metrics, showing that quantum variants can capture alternative similarity structures, although they do not consistently outperform classical methods.

Despite these advances, current QML approaches remain limited by the use of quantum components as feature transformers rather than integrated learning modules.

\section{Methodology}

\subsection{Overview}

\begin{figure}[H]
  \centering
  \includegraphics[width=0.95\columnwidth]{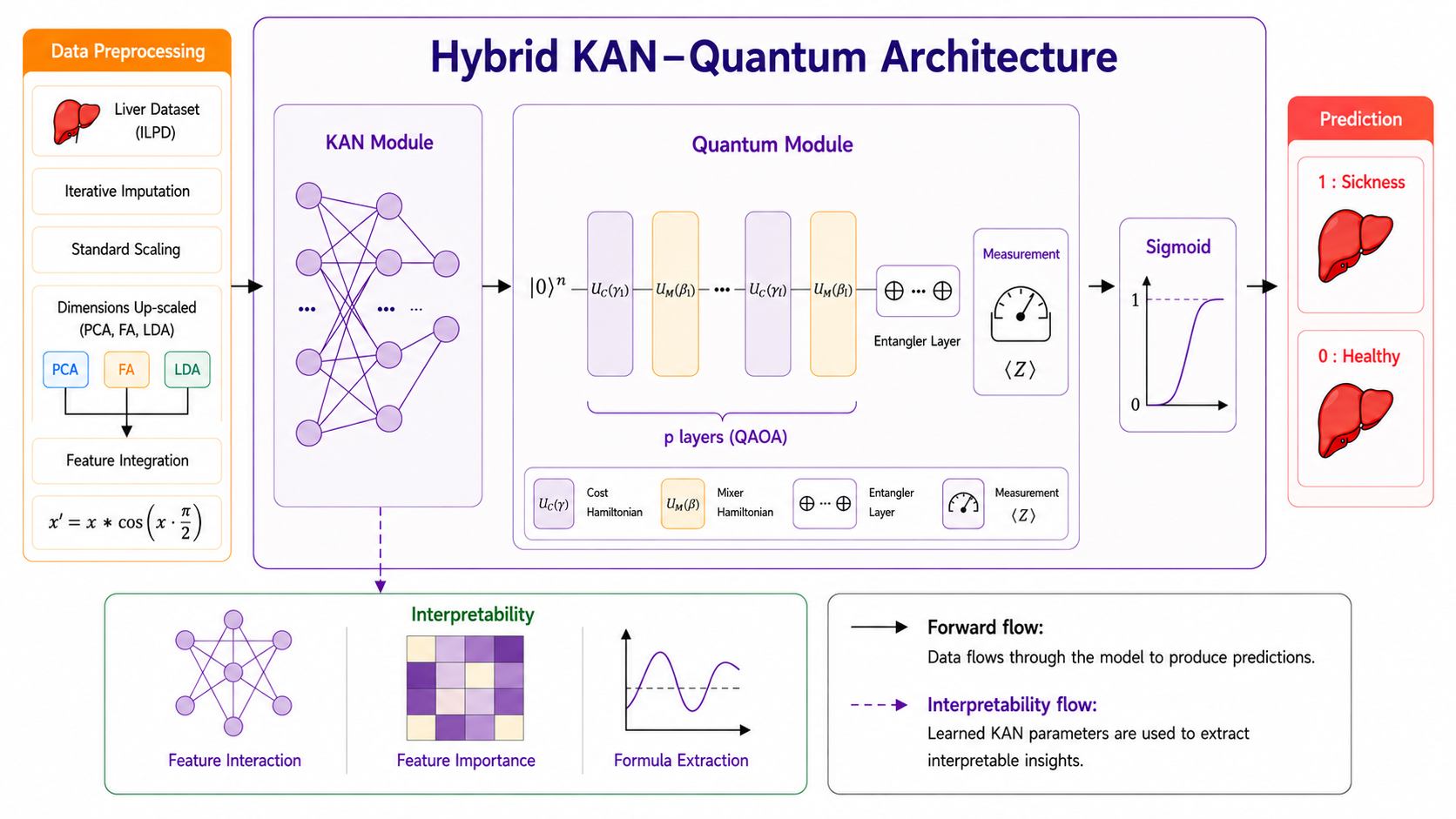}
  \caption{Overall Architecture of the Proposed Hybrid KAN–Quantum Framework.}
  \label{fig:quantum-kan-architecture}
\end{figure}

Figure~\ref{fig:quantum-kan-architecture} illustrates the overall workflow of the proposed hybrid KAN--quantum framework. The pipeline consists of three main stages: data preprocessing, hybrid feature learning, and interpretability. 

Raw clinical features are first processed through imputation, normalization, dimensionality enhancement, and feature integration. The resulting representation is then passed to a KAN module for nonlinear feature learning, followed by a lightweight quantum module that captures complementary higher-order interactions. Finally, a sigmoid layer produces the classification output. In addition to prediction, the framework provides multiple interpretability mechanisms.

\subsection{Feature Augmentation and Transformation}

The ILPD dataset contains a limited number of features, which constrains representational capacity and contributes to the modest performance gains observed across prior studies employing SVMs, Random Forests, and ensemble methods~[\citenum{ref:amin2023}--\citenum{ref:azam2020},\citenum{ref:quadir2023}]. To address this, we adopt a feature augmentation strategy that expands the input space rather than reducing it.

\begin{figure}[H]
  \centering
  \includegraphics[width=0.9\columnwidth]{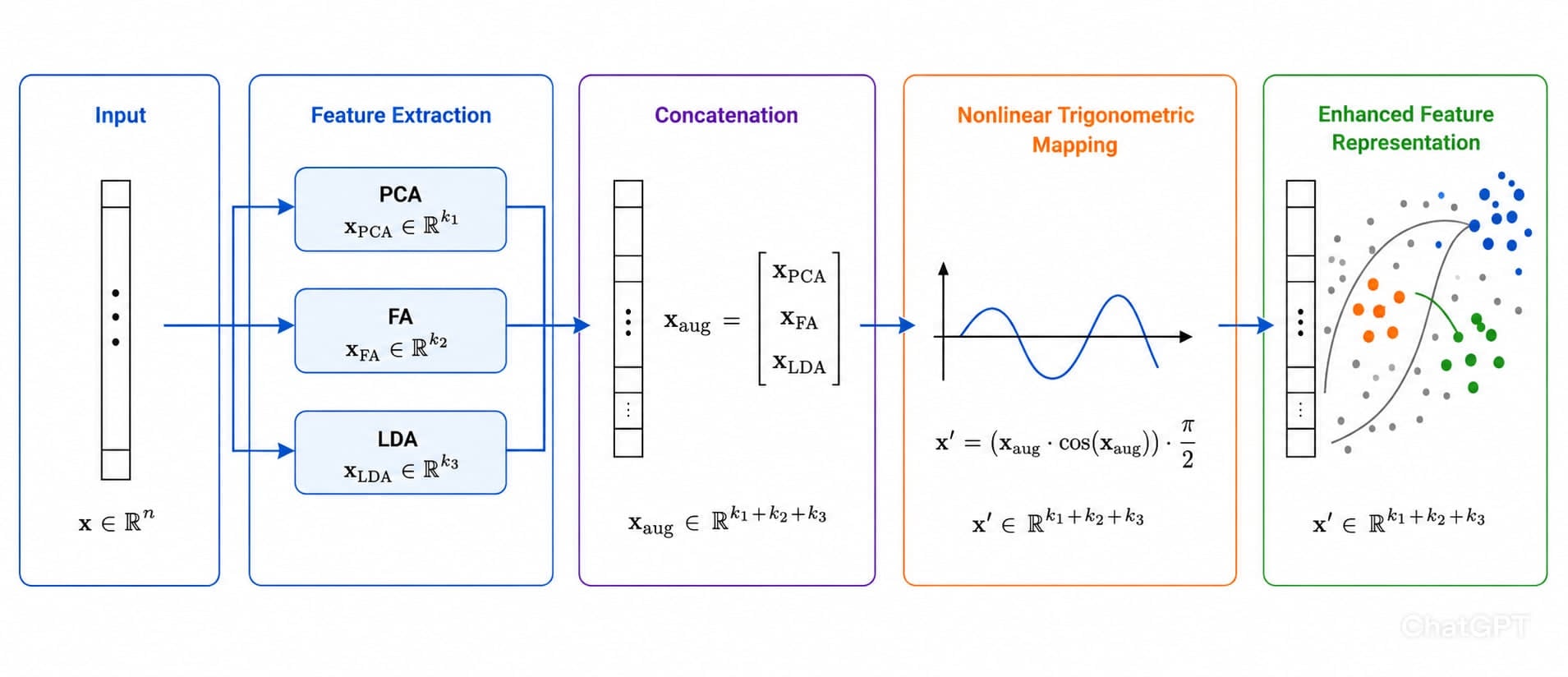}
  \caption{Feature augmentation and nonlinear transformation pipeline.}
  \label{fig:integrated-pipeline}
\end{figure}

PCA, Factor Analysis, and LDA are applied in parallel and their outputs concatenated, capturing variance, latent structure, and class-discriminative information respectively~\cite{ref:amin2023}. We further introduce a nonlinear trigonometric mapping $x' = (x \cdot \cos(x)) \cdot \frac{\pi}{2}$, which enriches the feature space with higher-order structure and facilitates accurate encoding of classical features as qubit rotation angles in the quantum domain. The full pipeline is illustrated in Fig.~\ref{fig:integrated-pipeline}.

\subsection{Hybrid KAN--Quantum Framework}

\begingroup
\setlength{\parskip}{0pt}
\setlength{\abovedisplayskip}{2pt}
\setlength{\belowdisplayskip}{2pt}
\setlength{\abovedisplayshortskip}{1pt}
\setlength{\belowdisplayshortskip}{1pt}

To address the dual challenges of high-dimensional input and limited quantum resources in the NISQ era, we propose a hybrid architecture that couples a Kolmogorov--Arnold Network (KAN) with a parameterized quantum circuit (PQC). The KAN first maps raw input $x \in \mathbb{R}^D$ to a compact latent vector $z \in \mathbb{R}^d$ ($d \ll D$), which is subsequently encoded into a quantum state for higher-order representation learning. This two-stage design decouples classical nonlinear compression from quantum feature interaction, allowing each component to operate within its computational regime.

The end-to-end pipeline can be summarized as:
\begin{equation}
  \hat{y} \;=\; \mathrm{sigmoid}\!\left(g\!\left(U(z)\,|0\rangle\right)\right),
  \quad z = f_{\mathrm{KAN}}(x),
  \label{eq:pipeline}
\end{equation}
where $U(z)$ is the parameterized encoding unitary and $g(\cdot)$ extracts the Pauli-$Z$ expectation values used as classical features for the final classifier.

\subsubsection{KAN-based Feature Learning}

The KAN module, illustrated in Fig.~\ref{fig:kan-quantum-diagram}, is structured as a composition of $L$ functional layers:
\begin{equation}
  z = f_L \circ f_{L-1} \circ \cdots \circ f_1(x),
  \label{eq:kan-composition}
\end{equation}
where each layer $f_\ell$ aggregates univariate spline-parameterized edge functions. Specifically, the output of node $j$ at depth $\ell$ is computed as:
\begin{equation}
  h_j^{(\ell)} = \sum_{i=1}^{m_{\ell-1}} \phi_{i,j}^{(\ell)}\!\left(h_i^{(\ell-1)}\right),
  \label{eq:node-aggregation}
\end{equation}
with $m_0 = D$ and $m_L = d$. Each $\phi_{i,j}^{(\ell)}$ is a learnable B-spline defined over a grid of control knots, enabling the network to approximate arbitrary smooth mappings on a per-edge basis. This edge-level parameterization provides substantially richer functional expressiveness than weight-scaled activations, while remaining interpretable through the shape of individual splines.

\begin{figure}[H]
  \centering
  \includegraphics[width=0.88\columnwidth, height=8cm]{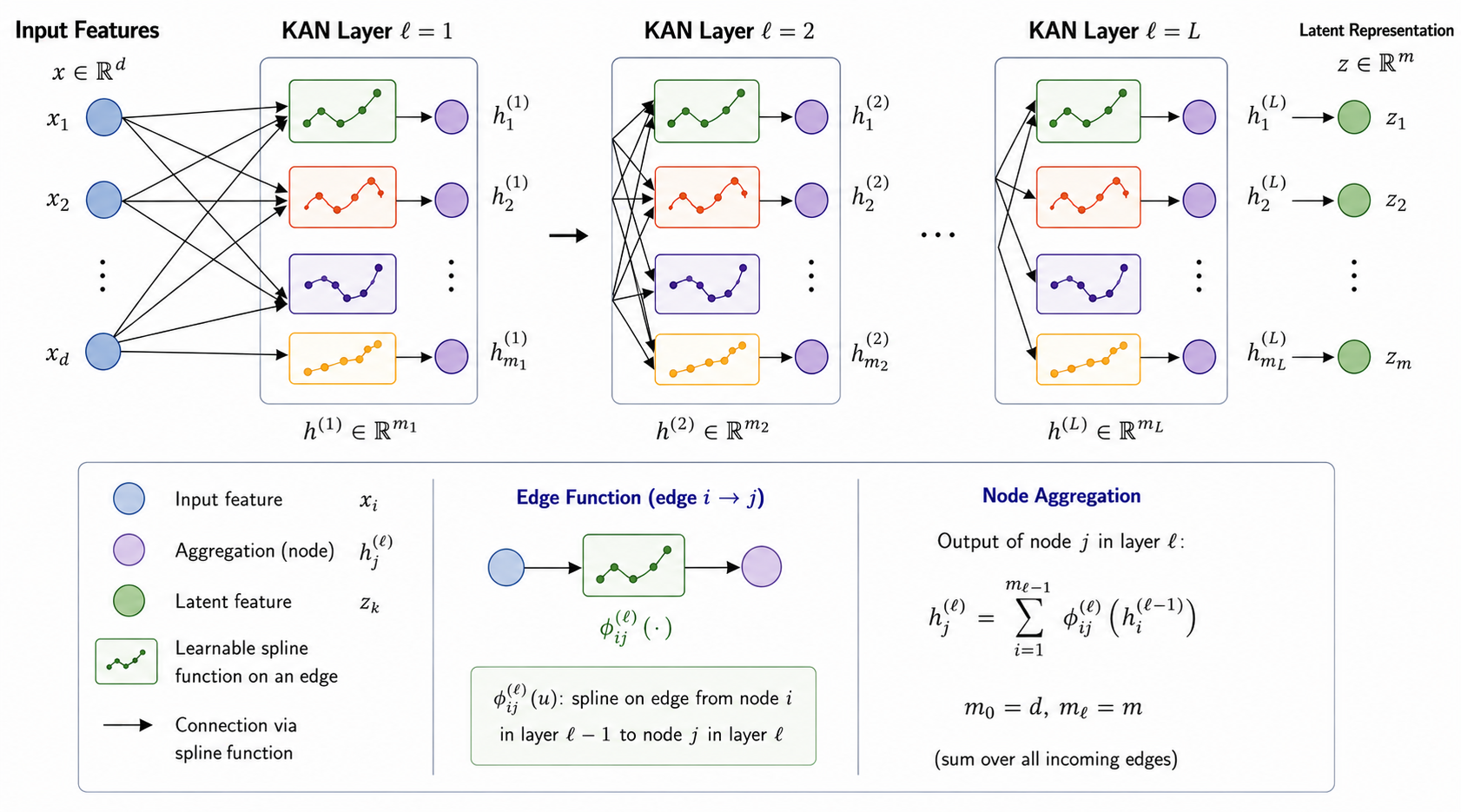}
  \caption{Architecture of the KAN module. Input features $x \in \mathbb{R}^d$ are processed through $L$ layers of learnable spline functions $\phi_{i,j}^{(\ell)}$ defined on control knots. Node outputs $h_j^{(\ell)}$ aggregate all incoming edge transformations (Eq.~\ref{eq:node-aggregation}), progressively compressing the representation to the latent vector $z \in \mathbb{R}^m$.}
  \label{fig:kan-quantum-diagram}
\end{figure}

Three architectural principles govern the module design. \textit{Progressive compression} schedules the layer widths $m_1 > m_2 > \cdots > m_L = d$ so that intermediate representations are preserved and information bottlenecks do not arise abruptly. \textit{Controlled depth} limits the number of layers to maintain generalization capacity on small datasets, where deep stacks risk overfitting. \textit{Moderate spline complexity}, realized through a modest number of grid points per edge, balances nonlinear flexibility with training stability.

\subsubsection{Quantum Feature Processing}

The latent vector $z \in \mathbb{R}^m$ is fed into the quantum module depicted in Fig.~\ref{fig:kan-quantum-diagram2}. Processing proceeds through three sequential stages.

\paragraph{Initialization.}
All $m$ qubits are placed into an equal superposition via Hadamard gates:
\begin{equation}
  |\psi_0\rangle = H^{\otimes m}|0\rangle^{\otimes m}.
  \label{eq:hadamard-init}
\end{equation}

\paragraph{QAOA Embedding.}
The circuit alternates between a Cost Hamiltonian $U_C(\gamma_\ell)$ and a Mixer Hamiltonian $U_M(\beta_\ell)$ for $L$ layers. The Cost Hamiltonian encodes the latent features and their pairwise interactions:
\begin{equation}
  H_C = \sum_i z_i Z_i + \sum_{i < j} J_{ij}\, Z_i Z_j,
  \label{eq:cost-hamiltonian}
\end{equation}
implemented via single-qubit phase rotations $R_Z(\gamma_{\ell,i}) = e^{-i\gamma_{\ell,i} Z_i}$ and two-qubit $Z_i Z_j$ interactions. The Mixer Hamiltonian $H_M = \sum_i X_i$ promotes exploration through bit-flip rotations $R_X(\beta_{\ell,i}) = e^{-i\beta_{\ell,i} X_i}$. All angles $\theta = \{\gamma_1, \beta_1, \ldots, \gamma_L, \beta_L\}$ are trainable parameters optimized via a classical optimizer.

\begin{figure}[H]
  \centering
  \includegraphics[width=0.95\columnwidth, height=8cm]{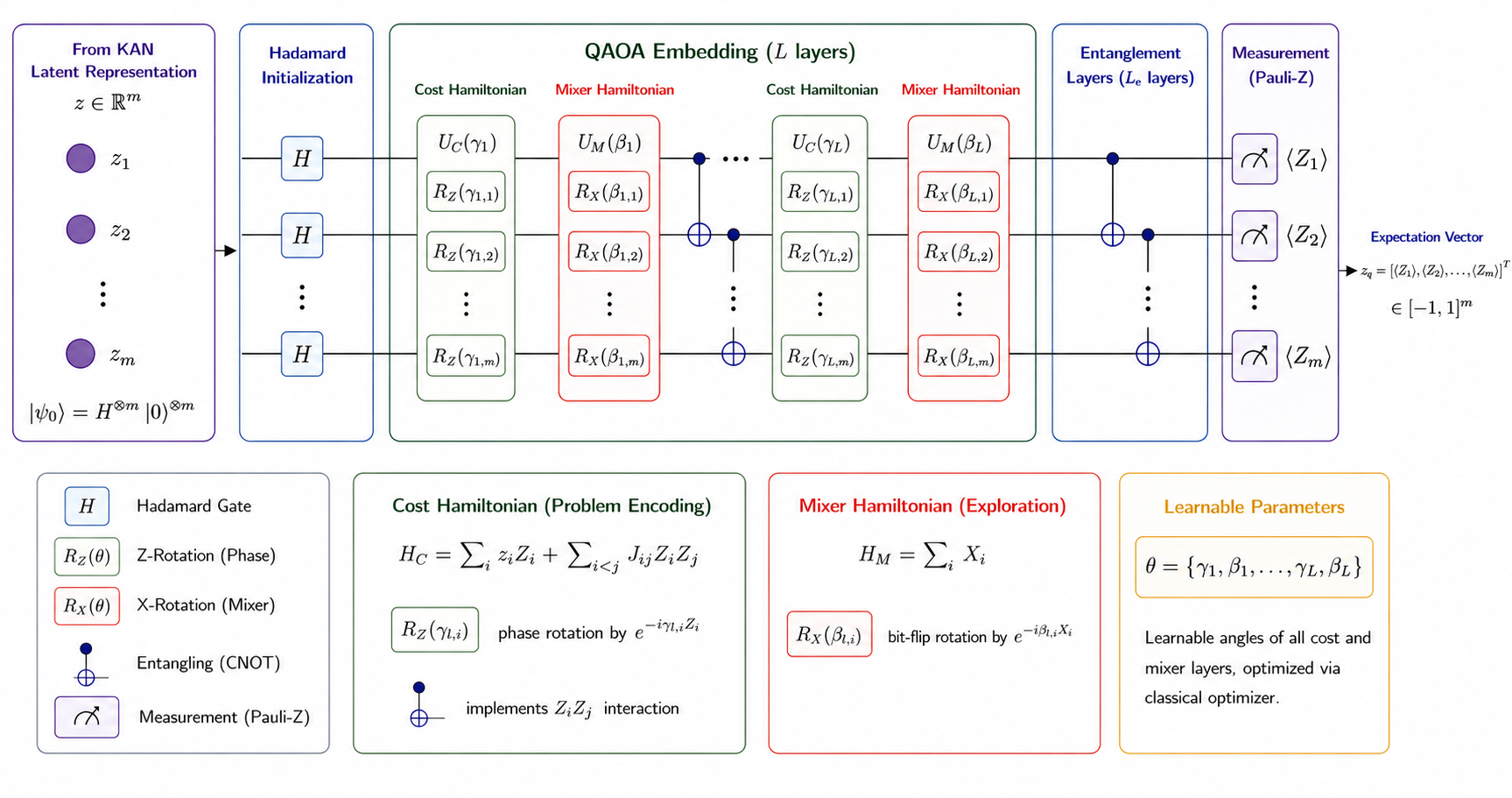}
  \caption{Architecture of the quantum module. The latent vector $z \in \mathbb{R}^m$ from the KAN is encoded via $L$ alternating Cost and Mixer Hamiltonian layers (QAOA Embedding), followed by $L_e$ CNOT-based entanglement layers. Pauli-$Z$ measurements yield the expectation vector $z_q \in [-1,1]^m$ (Eq.~\ref{eq:expectation-vector}), which drives the final sigmoid classifier.}
  \label{fig:kan-quantum-diagram2}
\end{figure}
\paragraph{Entanglement and Measurement.}
After the embedding layers, additional entanglement is introduced through CNOT gates arranged in $L_e$ entanglement layers to enhance inter-qubit correlations. Each qubit is then measured in the Pauli-$Z$ basis, yielding the expectation vector:
\begin{equation}
  z_q = \bigl[\langle Z_1 \rangle,\, \langle Z_2 \rangle,\, \ldots,\, \langle Z_m \rangle\bigr] \in [-1, 1]^m.
  \label{eq:expectation-vector}
\end{equation}
This bounded real-valued vector $z_q$ is passed to a sigmoid classifier to produce the final prediction $\hat{y}$.

\paragraph{Circuit configuration.}
The circuit is intentionally kept minimal, typically $2$--$3$ qubits and $1$--$2$ entangle layers, consistent with NISQ constraints. Shallow circuits are less susceptible to noise, and experiments show that increasing circuit depth or qubit count provides no significant performance gain. Since the KAN module already produces a compact and informative embedding, the quantum stage only needs a lightweight structure to capture residual higher-order interactions.

By pairing expressive spline-based feature compression with a resource-efficient quantum circuit, the proposed framework achieves competitive representational capacity under the computational constraints characteristic of near-term quantum devices.

\endgroup

\subsection{Handling Class Imbalance via Class Weight Smoothing}

To address class imbalance we adopt a class weight smoothing strategy based on the Effective Number of Samples (ENS) formulation~\cite{ref:cui2019}:

\begin{equation}
w_c = \frac{1 - \beta}{1 - \beta^{n_c}}, \quad c \in \{0,1\}
\label{eq:ens-weight}
\end{equation}

where $n_c$ is the number of samples in class $c$ and $\beta$ controls smoothing intensity. The raw weights are further refined via power scaling and Keras-style smoothing:

\begin{equation}
w_c \leftarrow w_c^\gamma, \quad 
w_c \leftarrow 1 + \alpha_{\mathrm{smooth}} \cdot (w_c - 1)
\label{eq:weight-refine}
\end{equation}

followed by normalization to unit mean, yielding balanced gradients without imposing excessive penalties on the majority class. We further incorporate Focal Loss (FL)~\cite{ref:focal2017} to emphasize hard examples and enhance discriminative learning within the KAN-derived embedding space.

\section{Experiment}

\subsection{Indian Liver Patient Dataset}
We use the ILPD from the UCI Machine Learning Repository~\cite{ref:ilpd2011}. The dataset contains 583 records with 10 clinical input features and one binary target label indicating liver disease status.

ILPD is moderately sized and clinically meaningful, but it presents two practical challenges: class imbalance (416 positive vs. 167 negative cases) and heterogeneous feature scales.

\subsection{Results}
\textbf{Baselines and metrics.} We benchmark our method against recent baseline approaches using the same dataset and evaluation protocol [\citenum{ref:alyasin2024},\citenum{ref:amin2023}, \citenum{ref:qml2025},\citenum{ref:kumarthakur2021},\citenum{ref:quadir2023},\citenum{ref:safriandono2024}]. For evaluation, we report Accuracy, Precision, Recall, F1-score to assess classification performance, robustness, and discrimination capability.

To ensure a fair comparison, we reproduced all reported methods under a strict leakage-free protocol. All preprocessing steps were fit only on the training split and then applied to validation and test splits, while the original ILPD class imbalance was preserved without synthetic resampling. As shown in \hyperref[tab:result-comparison]{Table~\ref*{tab:result-comparison}}, this controlled evaluation provides a reliable benchmark for assessing the proposed hybrid KAN--quantum framework.

\begin{table}[H]
\caption{Performance comparison under a unified evaluation protocol.}
\label{tab:result-comparison}
\centering
\resizebox{0.86\columnwidth}{!}{%
\begin{tabular}{llcccc}
\toprule
Author & Model & Acc. & Prec. & Rec. & F1 \\
\midrule
Quadir et al.~\cite{ref:quadir2023} & Extra Trees & 0.7402 & 0.7647 & 0.9218 & 0.8354 \\
Amin et al.~\cite{ref:amin2023} & RF+PCA+LDA+FA & 0.7200 & 0.7710 & 0.8680 & 0.8160 \\
Alyasin \& Ata~\cite{ref:alyasin2024} & Hybrid baseline & 0.8003 & 0.8028 & 0.9572 & 0.8700 \\
Kumar \& Thakur~\cite{ref:kumarthakur2021} & V-NWFKNN & 0.8100 & 0.8073 & 0.9424 & 0.8800 \\
Kumar \& Thakur~\cite{ref:kumarthakur2021} & V-NWFKNN (TL-RUS) & 0.8173 & 0.7982 & 0.9601 & 0.8900 \\
Safriandono et al.~\cite{ref:safriandono2024} & LR-QFE & 0.7400 & 0.7500 & 0.9900 & 0.8500 \\
Donaire et al.~\cite{ref:qml2025} & QML-Liver & 0.7500 & 0.7500 & 0.9400 & 0.8343 \\
\textbf{Ours} & \textbf{KQML-Liver} & \textbf{0.8198} & \textbf{0.8020} & \textbf{1.0000} & \textbf{0.8901} \\
\bottomrule
\end{tabular}%
}
\end{table}

\begin{itemize}
  \item \textbf{Precision and accuracy} achieve 80\% and 82\%, which are relatively high and outperform most reproduced results.
  \item \textbf{Sensitivity (Recall):} The model achieves a \textbf{100\%} sensitivity, outperforming most reproduced results. This performance is obtained under a strictly data-leakage-free evaluation protocol, where the training and test splits preserve the original data distribution. Such a high sensitivity is particularly desirable in medical applications, where failing to identify positive cases can lead to serious clinical consequences.
\end{itemize}

\subsection{Model Transparency Analysis}
\setlength{\textfloatsep}{8pt}
\setlength{\floatsep}{6pt}
\setlength{\intextsep}{6pt}

A key advantage of the proposed framework lies in its ability to provide interpretable insights into clinical features and their interactions. The KAN-based architecture enables improved analysis of feature importance, model structure, and learned functional relationships.

In this study, interpretability is examined from three complementary perspectives: feature importance, symbolic formula extraction, and feature interaction analysis.

\subsubsection{Feature Importance.}
We first analyze feature importance to identify which clinical variables most strongly influence model predictions using three complementary techniques: perturbation analysis, gradient-based sensitivity, and intrinsic magnitude analysis.

Perturbation analysis measures the change in model output when a feature is modified or removed, while gradient-based sensitivity evaluates how responsive predictions are to input variations, with larger values indicating greater importance. In addition, leveraging the inherent interpretability of KANs, we examine the intrinsic magnitude of learned spline functions. Unlike conventional MLPs, KAN provides a direct measure of feature contribution through these amplitudes, where larger magnitudes indicate higher importance.

Table~\ref{tab:feature-importance} presents the importance scores, showing TB emerges as the most dominant contributor across all metrics, followed by A/G ratio and liver enzymes (SGOT, SGPT). While DB, as the second bilirubin-related feature, ranks lower (6th by perturbation score), bilirubin as a group (TB + DB) collectively represents a strong diagnostic signal, consistent with clinical understanding.

\begin{center}
\captionof{table}{Feature importance scores from perturbation, gradient, and magnitude}
\label{tab:feature-importance}
\scriptsize
\setlength{\tabcolsep}{4pt}
\begin{tabular}{lccc}
\toprule
Feature & Perturb & Gradient & Magnitude \\
\midrule
TB      & 0.1502 & 0.1444 & 0.1521 \\
A/G     & 0.1241 & 0.1076 & 0.1298 \\
SGOT    & 0.1048 & 0.0980 & 0.1124 \\
TP      & 0.1028 & 0.0884 & 0.0913 \\
SGPT    & 0.0835 & 0.0800 & 0.1057 \\
DB      & 0.0685 & 0.0825 & 0.0859 \\
Age     & 0.0575 & 0.0511 & 0.0614 \\
Gender  & 0.0385 & 0.0476 & 0.0492 \\
ALB     & 0.0326 & 0.0303 & 0.0348 \\
AlkPhos & 0.0170 & 0.0248 & 0.0261 \\
\bottomrule
\end{tabular}
\end{center}

\subsubsection{Symbolic Formula Extraction.}

A key advantage of KANs is the ability to extract an explicit symbolic expression approximating the learned function, providing a transparent analytical representation.

Beyond interpretability, the extracted formula offers practical deployment advantages. It requires no deep learning framework or stored network weights, incurs negligible storage overhead, and enables fast inference through simple arithmetic operations. As a result, clinicians can estimate disease probability with minimal computational resources, making the approach suitable for resource-constrained clinical environments.

\subsubsection{Feature Interaction Analysis.}

To further investigate model behavior, we analyze pairwise feature interactions by computing interaction scores derived from KAN intrinsic properties and visualizing the most influential combinations. These maps reveal how clinical variables jointly influence liver disease risk.

\begin{figure}[H]
  \centering
  \begin{minipage}[t]{0.44\columnwidth}
    \centering
    \includegraphics[width=\linewidth]{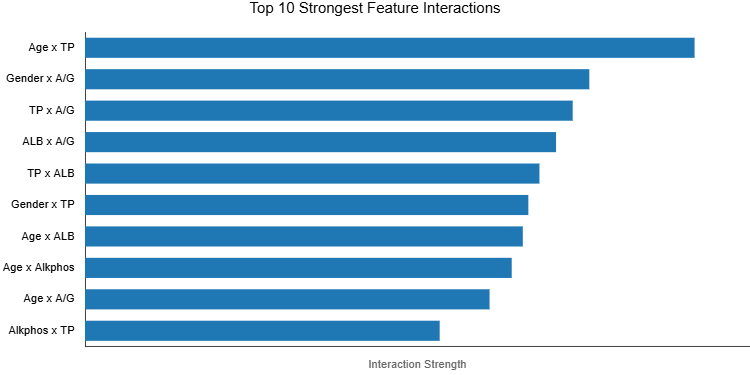}
    \\[-2pt]
    {\footnotesize (a) Pairwise interaction scores}
  \end{minipage}\hfill
  \begin{minipage}[t]{0.44\columnwidth}
    \centering
    \includegraphics[width=\linewidth]{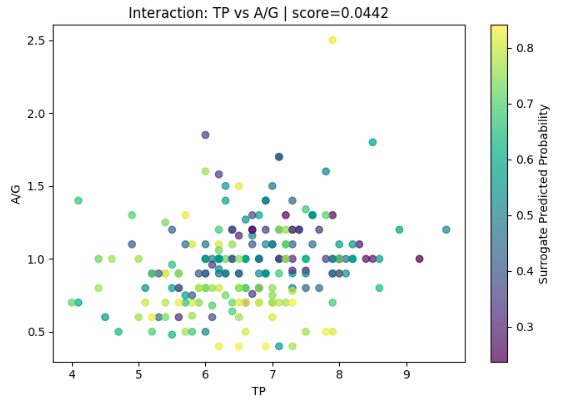}
    \\[-2pt]
    {\footnotesize (b) Interaction visualization}
  \end{minipage}
  \caption{Feature interaction analysis of the proposed model.}
  \label{fig:interaction-maps}
\end{figure}

The results highlight nonlinear dependencies between biomarkers that cannot be captured by simple linear models, as shown in Fig.~\ref{fig:interaction-maps}a. For example, certain combinations of A/G ratio and ALB, ALB, TP levels exhibit strong joint effects on prediction outcomes. Clinicians should pay close attention to these indicators when diagnosing the disease.

Based on these observations, clinically meaningful patterns can be identified. As shown in Fig.~\ref{fig:interaction-maps}b, lower A/G and TP levels are associated with higher disease risk, whereas higher values correspond to lower risk, consistent with established clinical evidence. Overall, the interaction analysis demonstrates that the proposed framework provides interpretable insights into feature relationships, supporting more reliable clinical decision-making.

\section{Conclusion and future work}

In this paper, we presented KQML-Liver, a hybrid KAN–quantum framework for liver disease classification that combines quantum feature learning with the interpretability of KANs. The proposed framework integrates rigorous preprocessing, leakage-free evaluation, and a lightweight architecture designed for imbalanced medical data. Experimental results on the ILPD dataset demonstrate strong predictive performance while maintaining interpretability through feature importance analysis, symbolic formula extraction, and feature interaction visualization.

However, further evaluation on datasets with different clinical characteristics, feature distributions, and class imbalance is still required to validate the model's generalizability. In future work, the proposed framework will be evaluated on larger and more challenging medical tabular datasets, such as the Arrhythmia~\cite{ref:guvenir1997}, to assess the consistency of its predictive performance and interpretability.

\begin{credits}
\subsubsection{\ackname} This research is supported by research funding from the Faculty of Information Technology, University of Science, Vietnam National University - Ho Chi Minh City.
\end{credits}

\section{References}
\begingroup
\renewcommand{\section}[2]{}
\bibliographystyle{splncs04}
\bibliography{references}

@misc{ref:alyasin2024,
  author = {E. I. Alyasin and O. Ata},
  title = {Enhancing the diagnosis of liver disease: Combining machine learning with the Indian liver patient dataset.},
  year = {2024},
  note = {In \textit{Proceedings of Fifth Doctoral Symposium on Computational Intelligence (DoSCI 2024)}, Vol. 3. DOI: \url{https://doi.org/10.1007/978-981-97-6036-7_19}.},
}

@misc{ref:guvenir1997,
  author = {H. Guvenir and B. Acar and H. Muderrisoglu and R. Quinlan},
  title = {Arrhythmia},
  year = {1997},
  note = {UCI Machine Learning Repository. DOI: \url{https://doi.org/10.24432/C5BS32}.},
}

@misc{ref:amin2023,
  author = {Ruhul Amin and Rubia Yasmin and Sabba Ruhi and Md. Habibur Rahman and Md. Shamim Reza},
  title = {Prediction of chronic liver disease patients using integrated projection based statistical feature extraction with machine learning algorithms.},
  year = {2023},
  note = {\textit{Informatics in Medicine Unlocked} 36 (2023), Article 101155. DOI: \url{https://doi.org/10.1016/j.imu.2022.101155}.},
}

@misc{ref:azam2020,
  author = {Md. Shafiul Azam and Aishe Rahman and S. M. Hasan Sazzad Iqbal and Md. Toukir Ahmed},
  title = {Prediction of liver diseases by using few machine learning based approaches.},
  year = {2020},
  note = {\textit{Australian Journal of Engineering and Innovative Technology} 2, 5 (2020), 85--90. DOI: \url{https://doi.org/10.34104/ajeit.020.085090}.},
}

@misc{ref:who-hepatitis,
  author = {{World Health Organization}},
  title = {WHO urges action on hepatitis announcing hepatitis D as carcinogenic},
  year = {2025},
  note = {World Health Organization News Release, July 28, 2025. Available at: https://www.who.int/news/item/28-07-2025-who-urges-action-on-hepatitis-announcing-hepatitis-d-as-carcinogenic (accessed: March 22, 2026)}
}

@misc{ref:liverdeath2023,
  author = {Sushma Cheemerla and Hareesh Balakrishnan},
  title = {Global Burden of Liver Disease: 2023 Update},
  year = {2023},
  note = {\textit{Journal of Clinical and Experimental Hepatology} 13, 1 (Jan-Feb 2023), 154--161. DOI: \url{https://doi.org/10.1016/j.jceh.2022.09.002}. PMID: 36643071.},
}

@misc{ref:qml2025,
  author = {Laura Mar{\'i}a Donaire and Gloria Ortega and Francisco Orts and Ester Mart{\'i}n Garz{\'o}n and Ernestas Filatovas},
  title = {Hybrid quantum-classical machine learning framework for liver disease detection},
  year = {2025},
  note = {{\it Engineering Applications of Artificial Intelligence}, 135 (2025), Article 109327. DOI: https://doi.org/10.1016/j.engappai.2025.109327},
}

@misc{ref:quadir2023,
  author = {Abdul Quadir Md and Sanika Kulkarni and Christy Jackson Joshua and Tejas Vaichole and Senthilkumar Mohan and Celestine Iwendi},
  title = {Enhanced preprocessing approach using ensemble machine learning algorithms for detecting liver disease.},
  year = {2023},
  note = {\textit{Biomedicines} 11, 2 (2023), Article 581. DOI: \url{https://doi.org/10.3390/biomedicines11020581}.},
}

@misc{ref:kumarthakur2021,
  author = {P. Kumar and R. S. Thakur},
  title = {Liver disorder detection using variable-neighbor weighted fuzzy K nearest neighbor approach.},
  year = {2021},
  note = {\textit{Multimedia Tools and Applications} 80 (2021), 16515--16535. DOI: \url{https://doi.org/10.1007/s11042-019-07978-3}.},
}

@misc{ref:raubitzek2024,
  author = {S. Raubitzek and S. Schrittwieser and A. Schatten and et al},
  title = {Quantum inspired kernel matrices: Exploring symmetry in machine learning.},
  year = {2024},
  note = {\textit{Physics Letters A} 525 (2024), 129895. DOI: \url{https://doi.org/10.1016/j.physleta.2024.129895}.},
}

@misc{ref:safriandono2024,
  author = {A. N. Safriandono and D. R. I. M. Setiadi and A. Dahlan and et al},
  title = {Analyzing quantum feature engineering and balancing strategies effect on liver disease classification.},
  year = {2024},
  note = {\textit{Journal of Future Artificial Intelligence and Technology} 1, 1 (2024). URL: \url{https://faith.futuretechsci.org/index.php/FAITH/article/view/12}.},
}

@misc{ref:bhaskaran2024,
  author = {P. Bhaskaran and S. Prasanna},
  title = {An accuracy analysis of classical and quantum enhanced K-nearest neighbor algorithm using canberra distance metric.},
  year = {2024},
  note = {\textit{Knowledge and Information Systems} (2024). DOI: \url{https://doi.org/10.1007/s10115-024-02229-w}.},
}

@misc{ref:kan2024,
  author = {Ziming Liu and Yixuan Wang and Sachin Vaidya and Fabian Ruehle and James Halverson and Marin Solja\v{c}i\'{c} and Thomas Y. Hou and Max Tegmark},
  title = {KAN: Kolmogorov-Arnold Networks.},
  year = {2024},
  note = {\textit{arXiv preprint} arXiv:2404.19756 (2024). DOI: \url{https://doi.org/10.48550/arXiv.2404.19756}.},
}

@misc{ref:focal2017,
  author = {Tsung-Yi Lin and Priya Goyal and Ross Girshick and Kaiming He and Piotr Doll\\'ar},
  title = {Focal loss for dense object detection.},
  year = {2017},
  note = {In \textit{Proceedings of the IEEE International Conference on Computer Vision (ICCV 2017)}, 2980--2988. DOI: \url{https://doi.org/10.1109/ICCV.2017.324}.},
}

@misc{ref:cui2019,
  author = {Yin Cui and Menglin Jia and Tsung-Yi Lin and Yang Song and Serge Belongie},
  title = {Class-balanced loss based on effective number of samples.},
  year = {2019},
  note = {In \textit{Proceedings of the IEEE/CVF Conference on Computer Vision and Pattern Recognition (CVPR 2019)}, 9268--9277. DOI: \url{https://doi.org/10.1109/CVPR.2019.00949}.},
}

@misc{ref:ilpd2011,
  author = {B. Ramana and V. V. Prasad and K. S. Babu},
  title = {ILPD (Indian Liver Patient Dataset).},
  year = {2011},
  note = {\textit{UCI Machine Learning Repository}. DOI: \url{https://doi.org/10.24432/C5D02C}.},
}

@misc{ref:sontakke2017,
  author = {S. Sontakke and J. Lohokare and R. Dani},
  title = {Diagnosis of liver diseases using machine learning},
  year = {2017},
  note = {In: 2017 International Conference on Emerging Trends \& Innovation in ICT (ICEI), IEEE, pp. 129--133. Available at: \url{https://ieeexplore.ieee.org/document/7977023}}
}
\endgroup

\end{document}